\documentclass[
]{ceurart}

\usepackage{listings}
\usepackage[utf8]{inputenc}
\usepackage[english]{babel}
\usepackage{graphicx, caption}
\usepackage[export]{adjustbox}
\usepackage{svg}
\usepackage{amssymb} 
\usepackage{amsthm} 
\usepackage[onehalfspacing]{setspace}
\usepackage{csquotes}

\begin{document}

\copyrightyear{2024}
\copyrightclause{Copyright for this paper by its authors.
  Use permitted under Creative Commons License Attribution 4.0
  International (CC BY 4.0).}
  
\conference{CHR 2024: Computational Humanities Research Conference, December 4–6, 2024, Aarhus, Denmark}

\title{Tracing the Development of the Virtual Particle Concept Using
Semantic Change Detection}

\author[1]{Michael Zichert}[%
email=m.zichert@tu-berlin.de,
orcid=0009-0007-8575-5750
]
\cormark[1]
  
\author[1]{Adrian Wüthrich}[%
email=adrian.wuethrich@tu-berlin.de,
orcid=0000-0002-6237-7327
]
\address[1]{History and Philosophy of Modern Science,
  Technische Universität Berlin, Germany}

\cortext[1]{Corresponding author.}

\begin{abstract}
Virtual particles are peculiar objects.
They figure prominently in much of theoretical and experimental
research in elementary particle physics.
But exactly what they are is far from obvious.
In particular, to what extent they should be considered \enquote{real}
remains a matter of controversy in philosophy of science.
Also their origin and development has only recently come
into focus of scholarship in the history of science.
In this study, we propose using the intriguing case of virtual particles to discuss the efficacy of Semantic Change Detection (SCD) based on contextualized word embeddings from a domain-adapted BERT model in studying specific scientific concepts.
We find that the SCD metrics align well with qualitative research insights in the history and philosophy of science, as well as with the results obtained from Dependency Parsing to determine the frequency and connotations of the term \enquote{virtual}.
Still,
the metrics of SCD provide additional insights over and above the
qualitative research and the Dependency Parsing.
Among other things,
the metrics suggest that the concept of the virtual particle became
more stable after 1950 but at the same time also more polysemous.

\end{abstract}

\begin{keywords}
 Semantic change detection
\sep digital conceptual history
\sep history and philosophy of science
\sep virtual particle
\end{keywords}

\maketitle


\section{Introduction}

Virtual particles have been important elements of particle physics
since long.
But despite their widespread use, the term \enquote{virtual particle}
holds different meanings and connotations within today's particle physics community,
and its historical origins and development have remained unclear.
Virtual particles are peculiar objects which may be considered responsible for the
fundamental interactions of matter and radiation.
In this sense, they have detectable and real effects.
However, they do not share the properties of real particles; for
instance, the mass and energy of a virtual particle does not stand in the same relation as
would be the case with a particle that is observed in the appropriate
detectors.
Virtual particles only ever occur in intermediate, unobservable phases
of decays or other processes involving elementary particles.
The precise meaning and interpretation of the term \enquote{virtual particle} has, therefore, been a topic for philosophical debate
\cite{Valente2011}.
Recent works by \citet{Ehberger2023} and \citet{Martinez2023} have
shed considerable light on the associated historical issues concerning
the origin and development of the virtual particle concept.
Additional studies on the conceptual shift due to Feynman diagrams and the associated calculation schemes have highlighted the relevance of virtual particles in the evolution of theoretical and experimental particle physics \cite{Kaiser2005, Wuthrich2010, Blum2017}.
While valuable, these studies are limited by their focus on carefully selected texts. 
Here, we aim to go beyond case studies and gain a more comprehensive view of the development of the concept of the virtual particle by analyzing a large dataset over an extended period of time.

To achieve this, we combine conceptual history with computational methods, an approach also referred to as \textit{digital Begriffsgeschichte} \cite{Wevers2020}.
First, we adapt our BERT model to the domain-specific language of our large corpus of physics texts and extract contextualized word embeddings for all occurrences of the term \enquote{virtual}.  
These word embeddings can then be used to employ Semantic Change Detection (SCD), which aims to identify, interpret and assess shifts in lexical meaning over time using computational techniques.
SCD has emerged as a distinct research field in recent years supported by multiple survey studies \cite[e.g.,][]{Periti2024, Tahmasebi2021}.
While most studies focus on the technical implementation of SCD, there have also been calls for further evaluation of the methods through in-depth case studies backed by qualitative analysis \cite{Periti2024a, Kutuzov2022}. 
We hope to provide such a case study with this paper.
To this end, we employ various SCD metrics to trace the origin, usage, and evolution of the concept of the virtual particle from a historical perspective, with special focus on the change in dominant meaning of the term \enquote{virtual} as well as its degree of polysemy, i.e., the coexistence of multiple meanings for a single word form.
For instance, the meaning of \enquote{virtual} in the context of \enquote{reality} differs from its meaning in the context of \enquote{particle.}
In order to enable a thorough evaluation of our results, we also use Dependency Parsing, thereby gaining a deeper understanding of the observed semantic shifts.\footnote{The code used in this study is available at \url{https://github.com/mZichert/scd_vp}. Due to copyright restrictions, the dataset and the domain-adapted BERT model used in the study are not available for public release.}

\section{Dataset}

\subsection{Physical Review corpus}

Our dataset consists of a large number of scientific articles from eight journals of the \textit{Physical Review}-family. 
The corpus spans from the introduction of the concept of virtuality in quantum physics in 1924 up to 2022, the latest complete year available for analysis, making it well-suited for studying the history of the virtual particle.
The \textit{PR}-journals are highly influential in the field of physics \cite{Bollen2006} and qualitative investigations \cite{Ehberger2023, Martinez2023} confirm their pivotal role in the emergence and establishment of the virtual particle concept, with several key articles on the topic published in these journals \cite[e.g.,][]{Bethe1936, Feynman1949a, Dyson1949}.
Through an agreement between our research project and the American Physical Society (APS), we have access to all normally restricted full texts, metadata, and citation data from this period \cite{AmericanPhysicalSociety}.
We include eight relevant journals into our analysis: \textit{PR - Series II} (all of physics until 1969), \textit{Review of Modern Physics} (long review articles with broad disciplinary scope, since 1929), \textit{PR - Letters} (short articles with high impact and broad disciplinary scope, since 1958), \textit{PR - A} (covering atomic, molecular and optical physics, since 1970), \textit{PR - B} (condensed matter and materials physics, since 1970), \textit{PR - C} (nuclear physics, since 1970), \textit{PR - D} (particle physics, field theory, gravitation, and cosmology, since 1970), and \textit{PR - E} (statistical, nonlinear, biological and soft matter physics, since 1993).
To focus on long-term trends, newer journals are excluded from the analysis.

The dataset's substantial size, comprising nearly 700,000 articles, makes it well-suited for extensive analysis using computational methods.
However, it also presents notable limitations, particularly concerning the early development of the concept. 
As a primarily US-based source written exclusively in English, significant developments from other regions are not captured. 
For instance, the center of the old quantum theory in the 1920s and early 1930s was in Central Europe, particularly in German-speaking countries, the Netherlands, and Denmark.
Since they also published mostly in German journals, most of their works are not a direct part of this study.
Another issue is the relatively small number of articles in the corpus published before 1950 (approximately 12,000 articles or just under 2 percent). 
For a more comprehensive analysis of the early phase of the concept using, it would be necessary to incorporate additional text sources.
 
\subsection{Data preprocessing}

Analyzing articles in the entire corpus using word embeddings is impractical due to scalability issues. 
Instead, we first identify articles potentially relevant to the concept of the virtual particle through a keyword search for \enquote{virtual} in the full texts, abstracts, and titles.
Approximately half of the full texts are available as digitized and OCR-processed PDF files (331,210 entries before 2004), while the other half are in native digital XML format (329,880 entries from 2004 onwards). 
For processing the PDF-files we use GROBID\footnote{GROBID stands for GeneRation Of BIbliographic Data (\url{https://grobid.readthedocs.io/en/latest/}).}, which allows parsing and restructuring of scientific publications in PDF format into uniformly TEI-formatted\footnote{\url{https://tei-c.org/}} XML files. 
To catch common OCR-errors prevalent the PDF-extracted text data, we apply some basic cleaning steps like removing special characters etc. 
Subsequently, citations and mathematical formulas are also removed from the text. 
While the formulas used likely reflect significant developments in the conceptualization of the virtual particle, there are currently no established tools for the content analysis of mathematical formulas in the context of conceptual history and the history of science.\footnote{We consider this an important open problem in semantic change detection in scientific texts.
Also, it is hard to estimate the impact of the omission of mathematical formulas.
On the one hand,
the symbols used in the formulas are usually explained in the surrounding text
(which we do take into account).
On the other hand,
different mathematical formulas may describe different virtual entities (particles, states, processes etc.) without clear indications of this in the surrounding text.
Moreover, as one of the anonymous reviewers pointed out, the frequency of formulas might have changed over time, which makes the omission potentially more or less impactful.
We did not control for this.}
Therefore, this work focuses on the analysis of linguistic text data. 

To ensure the efficient use of the BERT model the texts are segmented into sentences. 
For this task, we utilize the large language model of the Python natural language processing library SciSpaCy\footnote{\url{https://allenai.github.io/scispacy/}}, which has been trained on a large corpus of scientific texts (albeit in bio-medicine), making it suited for this purpose.
We also use the model for dependency parsing, where a sentence's syntactic structure is created by identifying how words are grammatically related through directed links.
This is particularly helpful for analyzing adjectives like \enquote{virtual}, as it allows for accurate identification of the associated nouns. 
We use these dependencies to evaluate and gain a deeper understanding of the observed semantic shifts.
Following \citet{Laicher2021}, we do not employ further preprocessing steps, such as lemmatization, as they do not seem to improve for SCD in English texts.  
After data preparation, our corpus consists 126,540 occurrences of \enquote{virtual}, spread across 41,786 articles.

\section{Methods}

\subsection{BERT and domain adaptation}
\label{bert}

For Semantic Change Detection using BERT\citep{Devlin2019}, fine-tuning for downstream tasks is unnecessary, as the focus is on the learned word representations, i.e., the contextualized word embeddings themselves.
Instead, BERT is adapted to the domain-specific language through re-training, a process known as domain adaptation. 
This involves reapplying Masked Language Modeling, enabling the model to learn the linguistic nuances and specialized terminology of the target domain. 
Domain adaptation is particularly crucial for this study, as the dataset comprises highly specialized scientific texts in physics. 
At the time of conducting our analysis, no suitable large language models specifically trained on general physics text data were available.
However, there are two models trained on specific sub-domains of physics: astroBERT\footnote{\url{https://huggingface.co/adsabs/astroBERT}} for astrophysics \cite{Grezes2021} and Astro-HEP-BERT\footnote{\url{https://huggingface.co/arnosimons/astro-hep-bert}} for astrophysics and (recent) high energy physics \cite{Simons2024}.\footnote{Recently, PhysBERT \cite{Hellert2024} was released, having been pre-trained on a large corpus of 1.2 million arXiv papers across various sub-fields of physics. While the model appears promising for our use case, it was released too late to be included in our study.}
For a comprehensive overview of scientific large language models, including those in the domain of physics, see \citet{Zhang2024}.

We therefore employ the BERT-base-uncased model\footnote{\url{https://huggingface.co/google-bert/bert-base-uncased}}, which features 12 attention layers and a hidden layer size of 768, and apply domain-adaption on our \enquote{virtual}-corpus.
We also re-trained and tested SciBERT \citep{Beltagy2019}\footnote{\url{https://huggingface.co/allenai/scibert_scivocab_uncased}}, which is primarily trained on scientific texts from biomedicine, but found that the re-trained BERT-base performs slightly better in terms of training and validation loss.
Regarding time-specific fine-tuning, we follow the findings from \citet{Martinc2020}, indicating that BERT's word embeddings are already well-suited to their temporal context due to their context-dependent nature.
For inference, the segmented sentences are fed into the model with a maximum sequence length of 512 tokens, and the sum of the last four layers is extracted for each token. 
For words comprising multiple subword tokens, the average embedding is stored. 
Given the contextual embeddings, each token occurrence results in one embedding vector. 
To reduce disk storage requirements, embeddings are saved only for meaningful words, excluding stop words, numbers, and special characters.

\subsection{Semantic Change Detection}
\label{scd}

\subsubsection{General workflow}

\renewcommand{\arraystretch}{1.5}
\begin{table}[!h]
\caption{Reference table of notations used in this paper.}
\centering
\begin{tabular}{|c|l|}
\hline
\textbf{Notation} & \textbf{Definition} \\ \hline
\(C\)           & Corpus \\ \hline
\(w\)           & Target word \\ \hline
\(t\)           & Time step in the investigation period $[1, \ldots, T]$ \\ \hline
\(s_{w}\)       & Semantic shift \(w\) \\ \hline
\(C_{w}^{t}\)   & Subcorpus containing \(w\) at \(t\)\\ \hline
\(\Phi_{w}^{t}\)  & Set of all embeddings of \(w\) in \(C_{w}^{t}\) \\ \hline
\(e_{w,i}^{t}\)   & $i$-th contextualized embedding of \(w\) in \(\Phi_{w}^{t}\) \\ \hline
\(\mu_{w}^{t}\)   & Word prototype of \(w\) for \(\Phi_{w}^{t}\) \\ \hline
\(\phi_{w,n}^{t}\)  & $n$-th cluster of embeddings of \(w\) in \(\Phi_{w}^{t}\) \\ \hline
\(P_{w}^{t}\)   & Cluster distribution of meaning clusters $\phi_{w,n}^{t}$ \\ \hline
\end{tabular}
\label{table:notations}
\end{table}

The basic procedure of Semantic Change Detection (SCD) can be outlined as follows:
Given a diachronic corpus of documents $C = \bigcup_{t=1}^{T} C_{w}^{t}$, where $C_{w}^{t}$ represents a subcorpus of documents at time $t$ within the overall investigation period $[1, \ldots, T]$ that contains the target word $w$.
The goal of SCD is to quantify the semantic shift $s_{w}$ for $w$ between two time-specific subcorpora $C_{w}^{t}$ and $C_{w}^{t'}$ or across the entire corpus.
There are two ways a semantic shift can manifest: firstly, as a change in the dominant meaning of a term, or secondly, as a change in the degree of its polysemy.
Both aspects will be analyzed in this study.
Specifically, for our purposes the target word is \enquote{virtual}, the documents comprise all the full texts plus abstracts of the \textit{PR}-corpus that contain \enquote{virtual}, and the time interval is one year. 

The generalized work-flow required for performing contextualized SCD can be split into three steps.
In the first step (\textbf{Embedding}), contextualized word embeddings are generated for each occurrence of the target word in the corpus using a large language model like BERT.
The set of all these embeddings in the time-specific subcorpus $C_{w}^{t}$ is expressed as $\Phi_{w}^{t} = \{e_{w,i}^{t}, \ldots, e_{w,I}^{t}\}$, where $e_{w,i}^{t}$ represents a contextualized word embedding in the subcorpus and $I$ denotes the number of all occurrences of $w$ in it. In the second step (\textbf{Aggregation}) the embeddings of a time period $\Phi_{w}^{t}$ are aggregated to represent the time-specific meanings of $w$.
Two types of representations are defined: 
\textit{Form-based} approaches examine the high-level properties of the target word per time period by looking directly at the dominant sense of a word or the degree of polysemy.
When considering the dominant meaning, word prototypes $\mu_{w}^{t}$ can be generated for each time interval representing the average of all embeddings in $\Phi_{w}^{t}$, thus providing an aggregated representation of the semantic properties of the target word in $C_{w}^{t}$.
When looking at polysemy at the high level, the aggregation step is usually skipped and the semantic shift of $w$ is measured by directly comparing the degree of polysemy in the time-specific set of embeddings $\Phi_{w}^{t}$ and $\Phi_{w}^{t'}$.
\textit{Sense-based} approaches, in contrast, attempt to first capture the different time-specific senses or meanings of the target word in $C_{w}^{t}$ using clustering methods.
Each time-specific meaning corresponds to a cluster of embeddings $\phi_{w,n}^{t}$ in the set of embeddings $\Phi_{w}^{t}$. 

We apply two clustering methods to identify meaning clusters. 
In \textbf{K-Means Clustering (KM)}, embeddings are organized into a predefined number of clusters by iteratively updating cluster centers until stable.
Determining the optimal number of clusters is challenging; automated methods like the silhouette coefficient often fail to identify the actual number of meaning clusters \cite{Martinc2020a}. 
Therefore, we set the number of clusters to $N = 10$ based on qualitative assessment.
\textbf{Affinity propagation (AP)} identifies exemplars among data points and forms clusters without the need to pre-specify their number by iteratively exchanging \enquote{messages} between data points to determine the clusters. 
However, the number of clusters often correlates with the number of input embeddings rather than actual meanings, potentially resulting in a large number of clusters \cite{Montariol2021}.  
Another drawback of AP is its high computational complexity of $O(n^2)$.
In our study, both clustering methods are applied to the entire corpus; however, it would also be feasible to employ time-specific clustering.
In order to make the clusters usable for SCD, we then calculate the probability distribution of the clusters, i.e., the cluster distribution $P_{w}^{t}$. 
The \textbf{cluster distribution} consists of the individual probabilities $p_{w,n}^{t}$, which indicate the frequency with which a specific embedding $e_{w,i}^{t}$ from the total set of embeddings $\Phi_{w}^t$ can be assigned to a particular cluster $\phi_{w,n}^{t}$. 
It is defined as follows:
\[	P_{w}^{t} = [p_{w,1}^{t}, p_{w,2}^{t}, \ldots, p_{w,N}^{t}] \textnormal{, where }  p_{w,n}^{t} = \frac{|\phi_{w,n}^{t}|}{|\Phi_{w}^{t}|}\cdot\]

Once the time-specific representations are identified, they can be compared over time in the final step (\textbf{Assessment}) to determine the extent of the semantic shift $s_w$. 
The methods used to quantify this shift, split into those measuring the semantic shift for polysemy and those for dominant meaning, will be introduced in the next chapters. 
Table \ref{table:notations} provides an overview of the notations used in this study.

\subsubsection{Polysemy}

We apply two methods to quantify the temporal development of a term's polysemy. 
The first method is \textbf{Shannon entropy} $H(P_{w}^{t})$, which utilizes the cluster distribution to describe the degree of uncertainty in the distribution of embeddings across meaning clusters within a given time period. 
Specifically, Shannon entropy quantifies the average amount of information needed to assign a particular embedding, i.e., an occurrence of the term \enquote{virtual}, to a specific cluster, i.e., a specific meaning of the term \enquote{virtual}.
A higher value of $H(P_{w}^{t})$ indicates a higher degree of polysemy, as there is greater uncertainty or variability in the cluster membership of the embeddings \cite{Baumann2023, Giulianelli2020}.
To ensure comparability of entropy values across different time periods, we use the normalized Shannon entropy $\eta(P_{w}^{t})$, which ranges from 0 to 1 and is defined as follows:

\[	\eta(P_{w}^t) = \frac{H(P_{w}^t)}{\log(N)} \textnormal{, where }  H(P_{w}^t) = -\sum_{n \in N} p_{w,n}^{t} \log(p_{w,n}^{t})\cdot\]

The second method, \textbf{Average Inner Distance (AID)}, utilizes the variance of the contextualized word embeddings $\Phi_{w}^{t}$, reflecting the degree of polysemy of $w$ in $C_{w}^{t}$.
In this approach, embeddings are not aggregated into meaning clusters or word prototypes.
Instead, the average distances between all possible pairs of embeddings within a single time period are calculated \cite{Periti2024}. 
This method is sometimes also referred to as self-similarity \cite{GariSoler2021}.
A higher AID value indicates greater polysemy of $w$ in $t$. 
We employ Euclidean distance, denoted in the formula as  $d(e_{w,i}^{t}, e_{w,j}^{t})$.
AID is defined as follows:

\[\mbox{AID}(\Phi_{w}^{t}) = \frac{1}{|\Phi_{w}^{t}|}\cdot \sum_{e_{w,i}^{t},e_{w,j}^{t}  \in \Phi_{w}^{t}, i<j} d(e_{w,i}^{t}, e_{w,j}^{t})\cdot\]

\subsubsection{Dominant meaning}

To assess the shift in dominant meaning in a form-based manner, Cosine Similarity (CS) can be used.
CS measures the alignment between the vectors of two word prototypes $\mu_{w}^{t}$ and $\mu_{w}^{t'}$ by calculating the dot product of the vectors divided by the product of their norms (lengths).
CS values range between -1 and 1, where a high value indicates vector alignment and a low value indicates opposition. 
We employ the variant \textbf{Inverted Cosine Similarity over Word Prototypes (PRT)}, which, according to \citet{Kutuzov2022}, is better suited for quantifying the extent of the semantic shift. 
PRT values are always greater than 1, where higher values signify a more pronounced shift. 
PRT is defined as follows:
\[\mbox{PRT}(\mu_{w}^{t}, \mu_{w}^{t'}) = \frac{1}{\mbox{CS}(\mu_{w}^{t}, \mu_{w}^{t'})} \textnormal{, where }  \mbox{CS}(\mu_{w}^{t}, \mu_{w}^{t'}) = \frac{\mu_{w}^{t} \cdot \mu_{w}^{t'}}{\left\|\mu_{w}^{t}\right\| \left\|\mu_{w}^{t'}\right\|}\cdot\]

The shift in dominant meaning can also be assessed using meaning clusters (sense-based) through the \textbf{Jensen-Shannon Divergence (JSD)}.
JSD, based on normalized Shannon entropy, measures the similarity between cluster distributions across different time periods.
This method considers not only the variation in the size of the clusters but also how the size of specific clusters across the different time periods changes \cite{Giulianelli2020}. 
A high JSD value indicates significantly different cluster distributions, suggesting pronounced semantic shifts.
Conversely, a low JSD value indicates relatively similar distributions, implying stability in the dominant meaning.
JSD is defined as follows:

\[ \mbox{JSD}(P_{w}^{t},P_{w}^{t'}) = H\left(\frac{1}{2}(P_{w}^{t}+P_{w}^{t'})\right) - \frac{1}{2} \left(H(P_{w}^{t})-H(P_{w}^{t'})\right)\cdot\]

\section{Results}

\subsection{Temporal development of \enquote{virtual}}

\begin{figure}[h]
\centering
\hspace{-0.0\linewidth}
\captionsetup{width=.9\linewidth}
\includegraphics[width=1.2\textwidth,center]{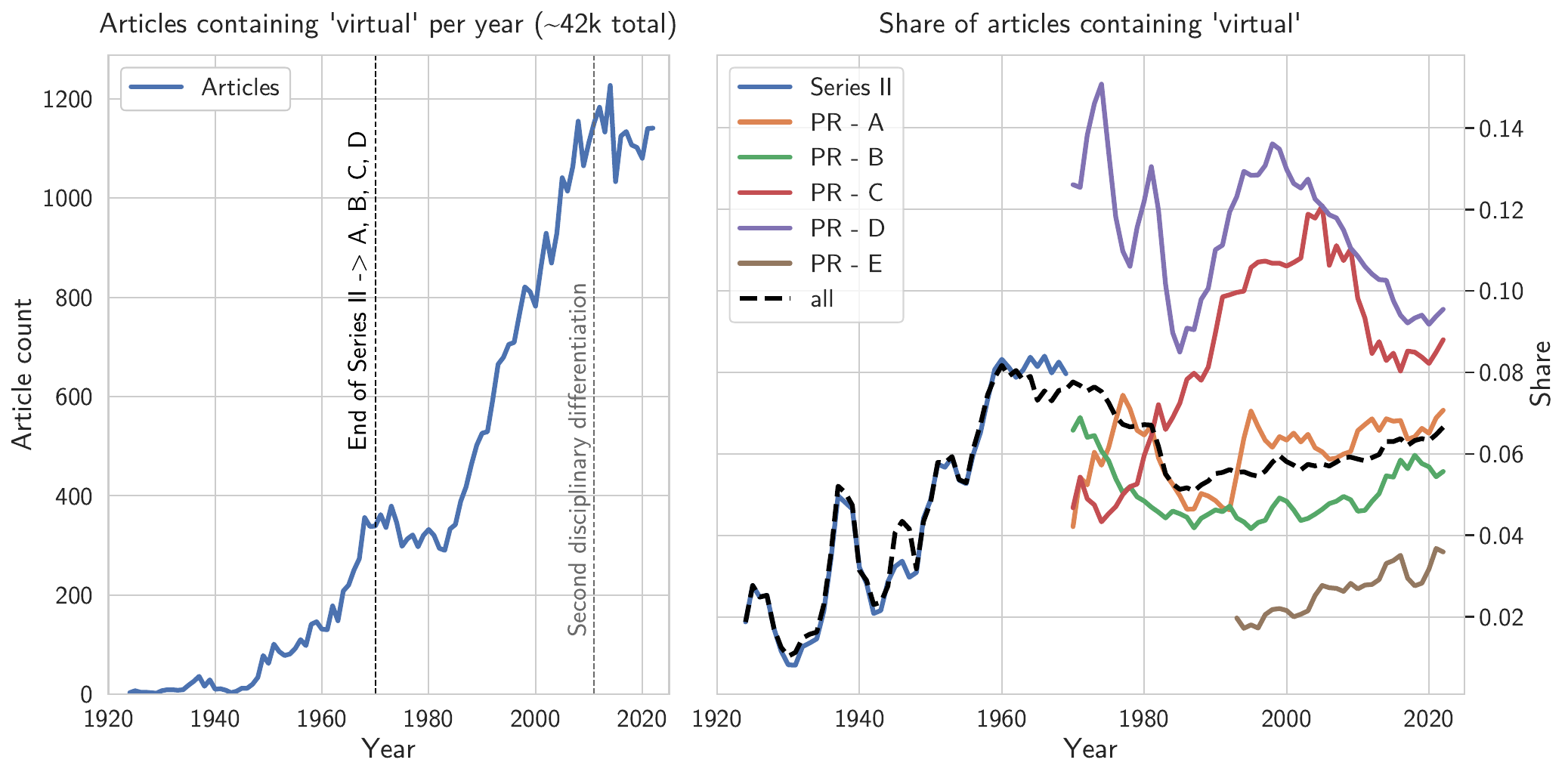}
\caption{Overview of the \textit{Physical Review} corpus: The figure displays the total number of published articles per year containing \enquote{virtual} for the entire corpus (on the left) and their proportion (rolling mean over 3 years) per journal (on the right). For clarity, the proportions in \textit{PR - Letters} and \textit{RMP} are not shown.}
\label{figure:corpus_overview}
\end{figure}

The first result of our study is the descriptive analysis of the \enquote{virtual} corpus in regards to the temporal development of the term.
Figure \ref{figure:corpus_overview} shows the number of published articles per year containing \enquote{virtual} for the entire corpus (left) and their proportion per journal (right). 
The dashed lines in the left figure indicate two key disciplinary differentiations in the \textit{PR} journals: the transition from \textit{Series II} to \textit{PR A - D} in 1970, and the introduction of new journals like \textit{PR - X} (2011) and \textit{PRX - Quantum} (2021). 
To focus on long-term trends, these newer journals are excluded from the analysis.
The decline in articles after 2010 is thus an artefact of the dataset and does not reflect overall trends in \textit{PR} publications or physics.
Notably, there is a low number of articles in the early phase of the study period, with only 384 publications in our corpus containing \enquote{virtual} before 1950, especially sparse before 1930 and during the war years (1942-1945). 
The exact number of articles, \enquote{virtual}-embeddings and cleaned tokens per year for the early phase can be found in the appendix (table \ref{table:corpus_counts}).
From 1950 onwards, the number of articles containing \enquote{virtual} increases steadily, with short periods of relative stagnation during the 1970s and 2010s, mirroring the broader increase in PR journal publications. 
Additional details on the total publication count per journal are available in the appendix (figure \ref{fig:num-virt}).
 
The average share of articles containing \enquote{virtual} across all journals, as depicted in the right figure, is 6.04 percent over the entire period. 
In the pre-Feynman era (before 1950), this percentage generally remains lower, except for two notable peaks.
In 1937, there is a temporary increase above 5 percent, driven by significant contributions from Bethe, Bacher, and Livingston in \textit{RMP} \cite{Bethe1936, Bethe1937, Livingston1937}.
The second peak in 1949 is best explained by Richard Feynman's groundbreaking articles and their reception.
For instance, with \textit{Space-Time Approach to Quantum Electrodynamics} \cite{Feynman1949a} -- published in \textit{PR - Series II} -- Feynman introduced his eponymous diagrams for representing and analyzing quantum electrodynamic processes, which contributed significantly to the establishment of the concept of the virtual particle.
In the same year, Freeman J.~Dyson's contributions, also published in \textit{Series II} \cite{Dyson1949, Dyson1949a}, further validated and established Feynman diagrams as a fundamental tool in quantum field theory (QFT) \cite{Ehberger2023, Wuthrich2010}.
Following the publications by Feynman and Dyson, the prevalence of \enquote{virtual} steadily increased, culminating in a peak during the 1960s and 1970s. 
This relatively high ratio of articles containing \enquote{virtual} may, at least in part, be due to the rise of an alternative to QFT: the so-called S-matrix theory \cite{Cushing1990}.
In this new theory, intermediate states were always on-shell such that it seems, at first sight, that \enquote{all talk of virtual particles was gone} \cite[p.~285]{Kaiser2005}. 
However, in other work by S-matrix theorists like Chew, Low, or Barut the virtual particle concept seems to take center stage, and even explicitly occurs in the title of one of their articles \cite{Chew1959a, Barut1962}.
Subsequently, from the 1970s onward, QFT emerged as the dominant theory, supported by its successful predictions and discoveries of fundamental particles such as quarks, W bosons, and Z bosons. 
Finally, by the early 1980s, the proportion of articles containing \enquote{virtual} starts to decline to approximately 5 percent, gradually rising again from the 1990s onward, albeit not returning to the levels observed during the earlier peak period.

Zooming in on the individual journals or disciplines respectively, articles containing \enquote{virtual} are notably prevalent in \textit{PR - D} (particle physics, field theory, gravitation, and cosmology) and \textit{PR - C} (nuclear physics).
Examination of arXiv classifications within \textit{PR - D} reveals that nearly 90 percent of these articles fall under high-energy physics.
The frequency of \enquote{virtual} in \textit{PR - D} peaks in the 1970s, 1990s and 2000s with drops in usage in between. 
Overall, it contributes approximately 27 percent of all articles containing the term \enquote{virtual} in the corpus, making it the largest source.
Nuclear physics (\textit{PR - C}) also features a significant percentage of articles containing \enquote{virtual}, comprising about 9 percent of the corpus. 
This aligns with recent research by Martinez on the origin of the notion of virtuality in modern physics \citep{Martinez2023}. 
The proportion of relevant articles in \textit{PR - C} increases steadily until the mid-1990s, plateaus until around 2010, and shows a recent decline.
The term is less prevalent in the remaining journals, which will not be discussed in detail here for the sake of brevity.
A table showing the top 5 journal-specific dependencies of \enquote{virtual} can be found in the appendix (table \ref{table:deps_journals_virtual}).

\subsection{Dominant meaning becomes more stable}
\label{sec:dom-stable}

\begin{figure}[h]
\centering
\hspace{-0.0\linewidth}
\captionsetup{width=.9\linewidth}
\includegraphics[width=1.2\textwidth,center]{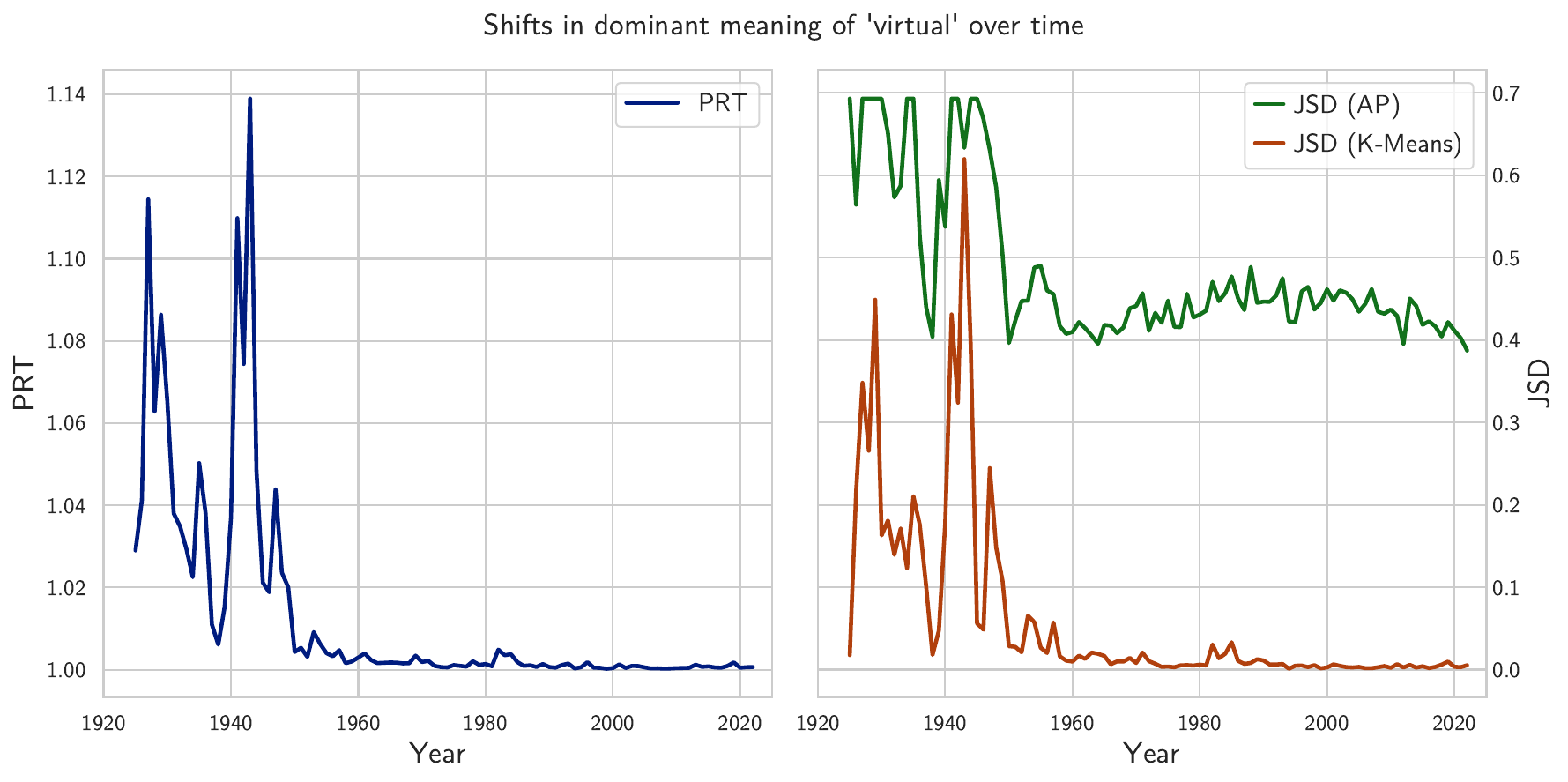}
\caption{Shifts in dominant meaning for \enquote{virtual}, using PRT (left) and JSD for K-Means and AP-clustering (right) in the entire \textit{PR}-corpus and over the entire investigation period.}
\label{figure:scd_domMeaning}
\end{figure}
  
One key finding of our study is that the dominant meaning of \enquote{virtual} becomes more stable over time. 
Figure \ref{figure:scd_domMeaning} presents the results of the SCD-calculations regarding the shifts in the dominant meaning throughout the entire investigation period.
The left graph displays the PRT-values for \enquote{virtual}, i.e., the inverted cosine similarity of the word prototypes for each year to preceding year. 
The right graph shows the JSD-values for both the K-Means-Clustering and the AP-Clustering.
Due to the computational expense of AP-Clustering, we randomly sampled approximately 25 percent of all embeddings, ensuring a minimum of 400 embeddings per year, where available. 

The resulting conceptual development of \enquote{virtual} can be divided into two distinct phases. 
The first period, up until the 1950s, is characterized by pronounced fluctuations, indicating repeated conceptual reorientation during the early development of the concept, with no firmly established or dominant meaning. 
This trend can be seen in all three metrics, although the values for JSD on the basis of AP-Clustering stabilizes at around 0.4.
Notably, peaks are observed in the late 1920s and early 1940s. 
Given the limited number of data points available for this period, it is important to emphasize that our results for this early period reflect general trends rather than individual peaks. 
To ensure the robustness of our results, we conduct permutation-based statistical tests, which are described in detail at the end of this chapter. 
From approximately 1950 onward, marking the beginning of the second phase, the dominant meaning begins to stabilize progressively, although a minor peak is observed in the early 1980s. 
This suggests the growing establishment of the concept of the virtual particle, following the outlined contributions of Feynman and Dyson.
Additional details on the shifts in dominant meaning in the discipline-specific journals can be found in the appendix (figure \ref{fig:domMeaning_journals}), indicating that the peak in the 1980s is mainly caused by a change in dominant meaning in \textit{PR - C} (nuclear physics).
We plan to conduct further research into the cause of this and other peaks.

Our findings regarding the stabilization of the dominant meaning of \enquote{virtual} are also supported by the time-specific dependencies, as shown in Table \ref{table:virtual_deps}. 
From the 1920s to the 1940s, \enquote{virtual} is most often associated with terms as diverse as \enquote{cathode}, \enquote{height}, \enquote{orbit}, \enquote{level}, and \enquote{oscillator}. 
In the 1940s, \enquote{virtual quanta} came into use, prominently featured in Feynman's first diagrams \citep{Feynman1949a}.  
With the onset of the post-Feynman era in the 1950s, \enquote{virtual photons} and \enquote{virtual states} become increasingly established as the dominant contexts.
Notably though, the concept of \enquote{virtual transition}, which Ehberger describes as essential for the concept's early development \citep{Ehberger2023}, only appears among the most frequent dependencies from 1960s on. 
From around 1990 onward, the dependency \enquote{correction} gains importance. 
These \enquote{virtual corrections} refer to parts of Feynman diagrams (or the corresponding mathematical expressions) involving the representation of a virtual particle. 
The increasing frequency of this use of \enquote{virtual} might be attributed
to an increasing interest in (and feasibility of) \enquote{higher order}
calculations and presicion measurements in various contexts,
the most prominent being the search for the Higgs boson at the Large Electron–Positron
Collider (LEP),
which was in use at CERN from 1989 to 2000,
the Tevatron (at Fermilab, 1983--2011), 
the planned Superconducting Super Collider (SSC, planned ca.~1983, cancelled in 1993),
and
at the Large Hadron
Collider
(LHC), which has been in use at CERN since 2009.%
\footnote{For an non-technical overview of higher order calculations,
  see \cite{Zanderighi2017}}
Nonetheless, \enquote{virtual photons} and \enquote{virtual states} remain the dominant contexts of use until the present, though less pronounced than in the 1960s and 1970s.

\begin{table}
\caption{Top 4 lemmatized dependencies of \enquote{virtual} per decade. The number in brackets represents the share of the dependency in all dependencies of the decade.}
\centering
\begin{tabular}{|c|c|c|c|c|}
\hline
Decade & Top 1 & Top 2 & Top 3 & Top 4 \\
\hline
1920 & cathode (23\%) & orbit (14\%) & oscillator (12\%) & radiation (7\%) \\
1930 & height (22\%) & level (18\%) & state (9\%) & oscillator (5\%) \\
1940 & height (13\%) & quanta (11\%) & level (8\%) & state (8\%) \\
1950 & photon (11\%) & state (10\%) & meson (9\%) & process (6\%) \\
1960 & photon (13\%) & state (12\%) & transition (5\%) & excitation (5\%) \\
1970 & photon (21\%) & state (11\%) & excitation (5\%) & transition (3\%) \\
1980 & photon (15\%) & state (11\%) & transition (4\%) & excitation (4\%) \\
1990 & photon (14\%) & state (8\%) & transition (3\%) & correction (3\%) \\
2000 & photon (14\%) & state (8\%) & correction (4\%) & excitation (3\%) \\
2010 & photon (12\%) & state (7\%) & correction (4\%) & process (3\%) \\
2020 & photon (14\%) & state (7\%) & correction (3\%) & orbital (2\%) \\
\hline
\end{tabular}
\label{table:virtual_deps}
\end{table}

The consistency of results across all three calculation methods, despite their different approaches, also notable: 
The values of PRT strongly correlate with those of JSD (Pearson coefficient for PRT and JSD - KM: 0.96, PRT and JSD - AP: 0.8), as well as the those of the two JSD metrics (0.77).
These high correlation values suggest that both clustering methods reliably identify the various meanings of \enquote{virtual}, indicating stable and meaningful results. 
To further ensure the robustness of our findings despite the relatively low frequency of \enquote{virtual} in the early years, we employ permutation-based statistical tests for the PRT-metric, following the approach outlined in \citet{Liu2021}.
Permutation tests can be used to assess whether the observed test statistic (i.e., the SCD-metrics) differs significantly from zero, therefore indicating a semantic shift between two time periods. 
These tests are particularly suitable for low-frequency data because they do not rely on large sample sizes or specific distributional assumptions; instead they generate the sampling distribution based on the available data itself. 
This is achieved through the random and repeated rearrangement of the \enquote{virtual}-embeddings across the two time periods by sampling without replacement and then recalculating the SCD-metric for each permutation.\footnote{We limit the number of permutations to a maximum of 100,000 per time interval, i.e. two subsequent years, to save computational resources.}
Following \citet{Liu2021}, we employ the Benjamini-Hochberg procedure to adjust the $p$-values for multiple comparisons, thereby limiting the false discovery rate. 
Applying this method to our data, we find that the semantic shifts for the dominant meaning of \enquote{virtual} based on PRT are significant for almost all time interval.
These findings support our conclusion regarding the general trend of the conceptual development while acknowledging variability in specific time periods.
A detailed exemplary figure illustrating the results of the permutation tests for PRT can be found in the appendix (figure \ref{fig:prt_perm_testing}).

\subsection{Polysemy increases}

\begin{figure}[h]
\centering
\hspace{-0.0\linewidth}
\captionsetup{width=.9\linewidth}
\includegraphics[width=1.2\textwidth,center]{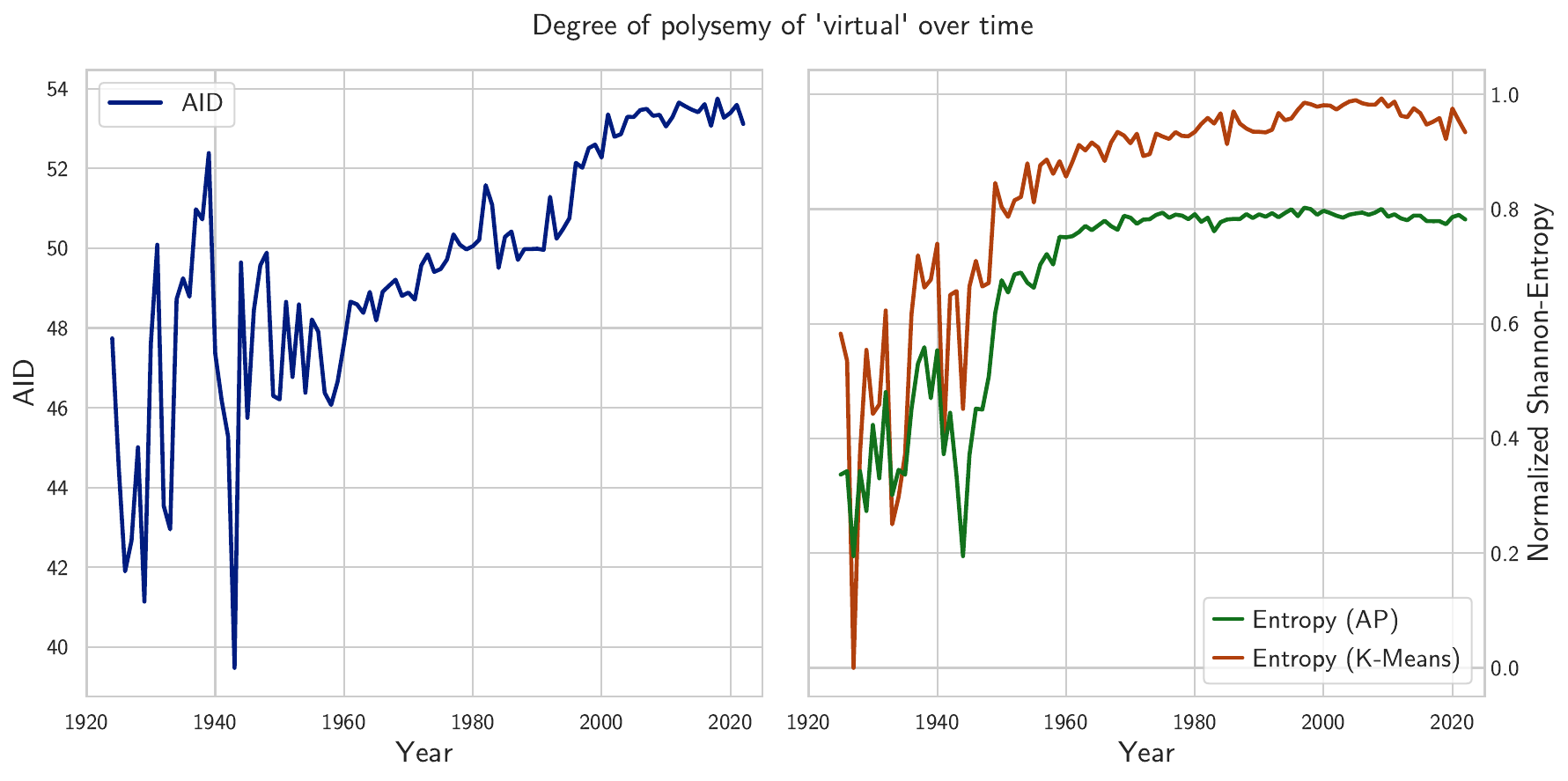}
\caption{Changing degree of polysemy for \enquote{virtual}, using AID (left) and normalized Shannon-Entropy for K-Means and AP-clustering (right) in the entire \textit{PR}-corpus and over the entire investigation period.}
\label{figure:scd_polysemy}
\end{figure}
  
The second key finding of our study is that the degree of polysemy of \enquote{virtual} increases.
That means that while the most dominant use is that in association with the aforementioned concepts, its usage in different meanings is also expanding.
Figure \ref{figure:scd_polysemy} presents the development of the degree of polysemy for \enquote{virtual} in the entire \textit{PR}-corpus and over the entire investigation period.
The left graph shows the AID-values, i.e., the average inner distances of all \enquote{virtual} embeddings in a given year. 
The values for the normalized Shannon-Entropy are displayed in the right graph, again for both the K-Means-Clustering and the AP-Clustering (with the same random sampling as described in Section~\ref{sec:dom-stable}).

Similar to the results regarding the dominant meaning, the degree of polysemy fluctuates significantly in the early phase of the concept. 
Notably, the values are particularly low in the mid to late 1920s and early 1940s.
These results are expected given the limited number of articles during these periods, as a small number of embeddings implies a correspondingly low number of different meanings.
From 1938 to 1940, however, the values for all calculation methods are particularly high.
A clear explanation for this spike is not immediately apparent, as neither the examination of the dependencies nor the shift in dominant meaning during these years provide insight. The described peaks in PRT and JSD occur several years later.
One possible explanation could be that few but very different embeddings cause the peak. 
While the correlation coefficients between the metrics are again high (0.64 for AID and Entropy (KM), 0.66 for AID and Entropy (AP), and 0.94 for Entropy (KM) and Entropy (AP)), suggesting stable results, we were, however, unable to identify a suitable method for statistical testing of polysemy.
Further research and qualitative assessment of the relevant papers is required and planned.
Consequently, our present analysis focuses, once again, on general trends rather than individual peaks.

From around 1950 or 1960, depending on the metric, the fluctuations become smaller and the degree of polysemy continues to steadily increase.
Notably, there is a brief spike in the early 1980s in the AID-values and another sharp increase in the 1990s, followed by a relative stabilization in recent years.
This increase in recent years is also reflected in the dependencies of \enquote{virtual} (table \ref{table:virtual_deps}), with the most frequent usage contexts becoming more evenly distributed from the 1990s compared to earlier decades.
This trend is supported by the introduction of the journal \textit{PR - E} in 1993, which is characterized by distinct usage contexts differing from those of other journals (see table \ref{table:deps_journals_virtual}).
The Shannon-Entropy based on both clustering methods remains consistently high, exceeding or maxing out at 0.8 from about the 1950s onward and reaching nearly maximum values around the 2000s in the case of K-Means. 
From 2010 onward, there is a small decrease in polysemy, possibly due to the second disciplinary differentiation leading to a slightly less varied usage of the term across the remaining journals.
The trends observed in discipline-specific journals generally align with the overall findings.
The details can be found in the appendix (figure \ref{fig:polysemy_journals}).

\section{Discussion}

We have used a large number of contextualized word embeddings to employ various Semantic Change Detection metrics in order to trace the diachronic development of the concept of the virtual particle. 
Our findings show that the dominant meaning of \enquote{virtual} becomes more stable over time while at the same time its degree of polysemy is increasing.
This development can be split into two periods: An initial phase characterized by repeated conceptual reorientation with no firmly established meaning yet, and a second phase marked by the growing consolidation of the dominant meaning in the sense of the virtual particle, following the seminal works of Richard Feynman and their reception around 1950.
Simultaneously, the degree of polysemy steadily increases throughout almost the entire investigation period and only recently seems to stabilize at a high level. 

While these two findings might seem contradictory at first, they can easily be reconciled. 
Simply put, the metrics for polysemy measure how spread out the word embeddings are in the vector space, while the metrics for dominant meaning measure where the relative majority of the embeddings lie and how this position changes from year to year. 
Our findings suggest that from the 1950s onward, the relative majority of the embeddings consistently centers around a usage in the sense of the virtual particle (especially virtual photons), while the overall usage of the term \enquote{virtual} diversifies, possibly due to its uses in different disciplines like those in \textit{PR - E}.

We have combined our SCD-based approach with evaluation via Dependency Parsing as well as qualitative assessment of the results.
We find that the observed semantic shifts are largely supported by recent work in the history of the virtual particle.
This is particularly true for the first period of the conceptual development, whereas SCD can be employed in a more heuristic manner for the still relatively under-researched second phase. 
For instance, we identified a notable and unexpected shift in dominant meaning in the 1980s, primarily driven by articles in nuclear physics (\textit{PR - C}). We plan to conduct further research into this peak as well as a more in-depth discussion of the relevance of our findings for the history and philosophy of physics.%
\footnote{%
Many further preliminary results are contained in M.Z.'s master thesis on the topic \citep{Zichert2023}.
Here, we focused on advocating a new method (Semantic Change Detection) for studying concepts in science.}
The complementary method of Dependency Parsing revealed that most of the semantic shifts coincide with significant changes in the most prominent dependencies at that time.
While Dependency Parsing may have been particularly effective in our case because \enquote{virtual}, the focus of our study, is an adjective, it could prove to be a valuable and resource-efficient evaluation method for broader use in SCD research.

\begin{acknowledgments}

This work was supported by the DFG Research Unit “The Epistemology of
the Large Hadron Collider” (Grant FOR 2063).
The members of the Unit provided valuable feedback at several stages
of this work.
Special thanks go to Robert Harlander, Jean-Philippe Martinez, Rebecka
Mähring, Arno Simons and Friedrich Steinle as well as three anonymous reviewers for their comments and
helpful suggestions. 
The work is based on M.Z.'s MSc thesis, which has been defended at the
University of Leipzig (Computational Humanities Research Group), and
was supervised by A.W. and Andreas Niekler.
We are also grateful to the American Physical Society for granting us
access to the relevant full texts and metadata.
\end{acknowledgments}

\bibliography{main}

\newpage
\appendix
\section*{Appendix}
\addcontentsline{toc}{section}{Anhang}
\renewcommand{\thesubsection}{\Alph{subsection}}

\subsection{Figures}

\begin{figure}[h!]
  \centering
  \includegraphics[width=1\textwidth,center]{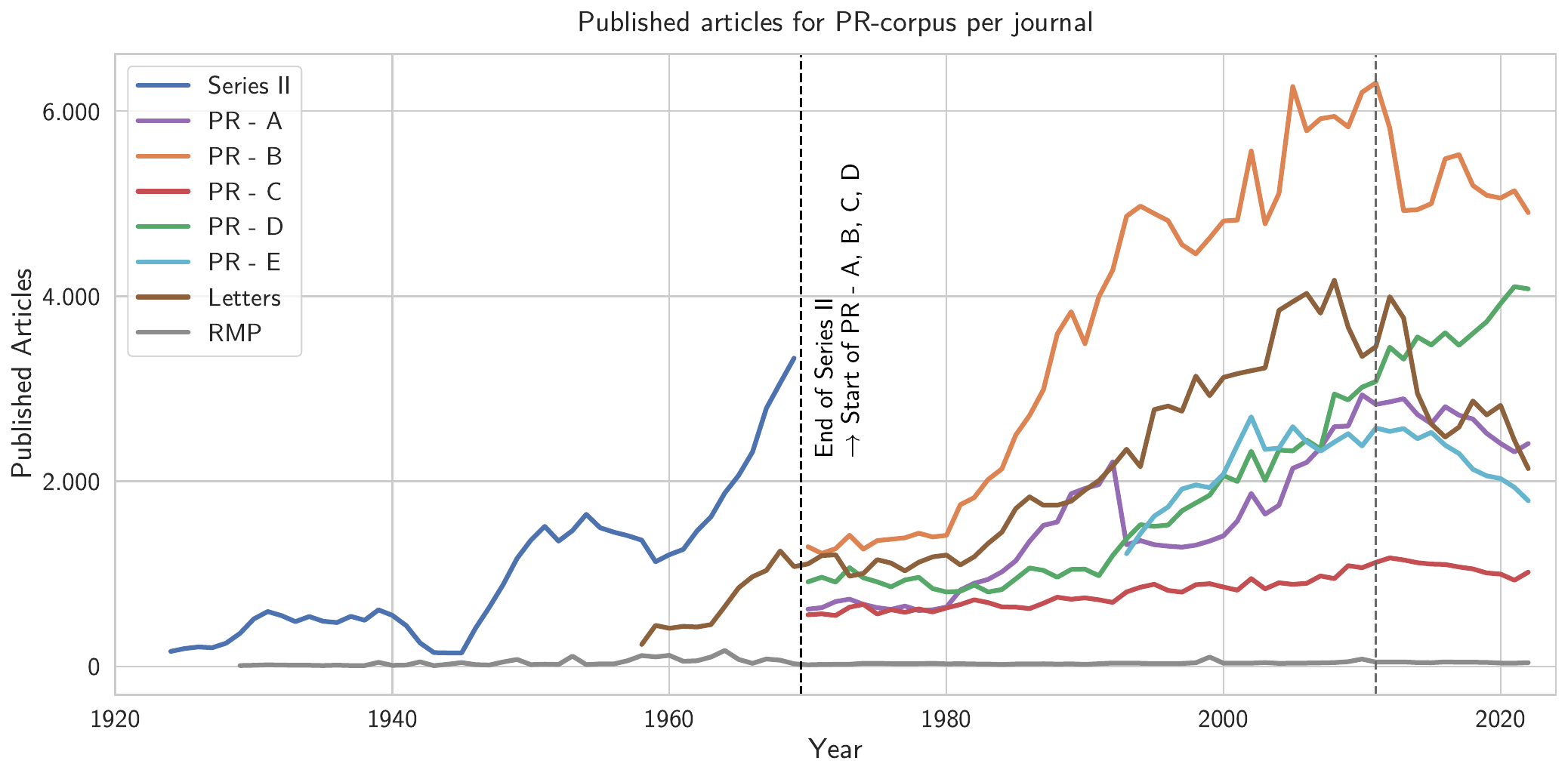}
  \caption{Number of published articles per year for each journal in the \textit{PR}-corpus. The first dashed line indicates the transition from \textit{Series II} to \textit{PR A - D}, while the second dashed line marks a subsequent disciplinary differentiation around 2010.}
  \label{fig:num-virt}
\end{figure}

\begin{figure}[h!]
  \centering
  \includegraphics[width=1\textwidth,center]{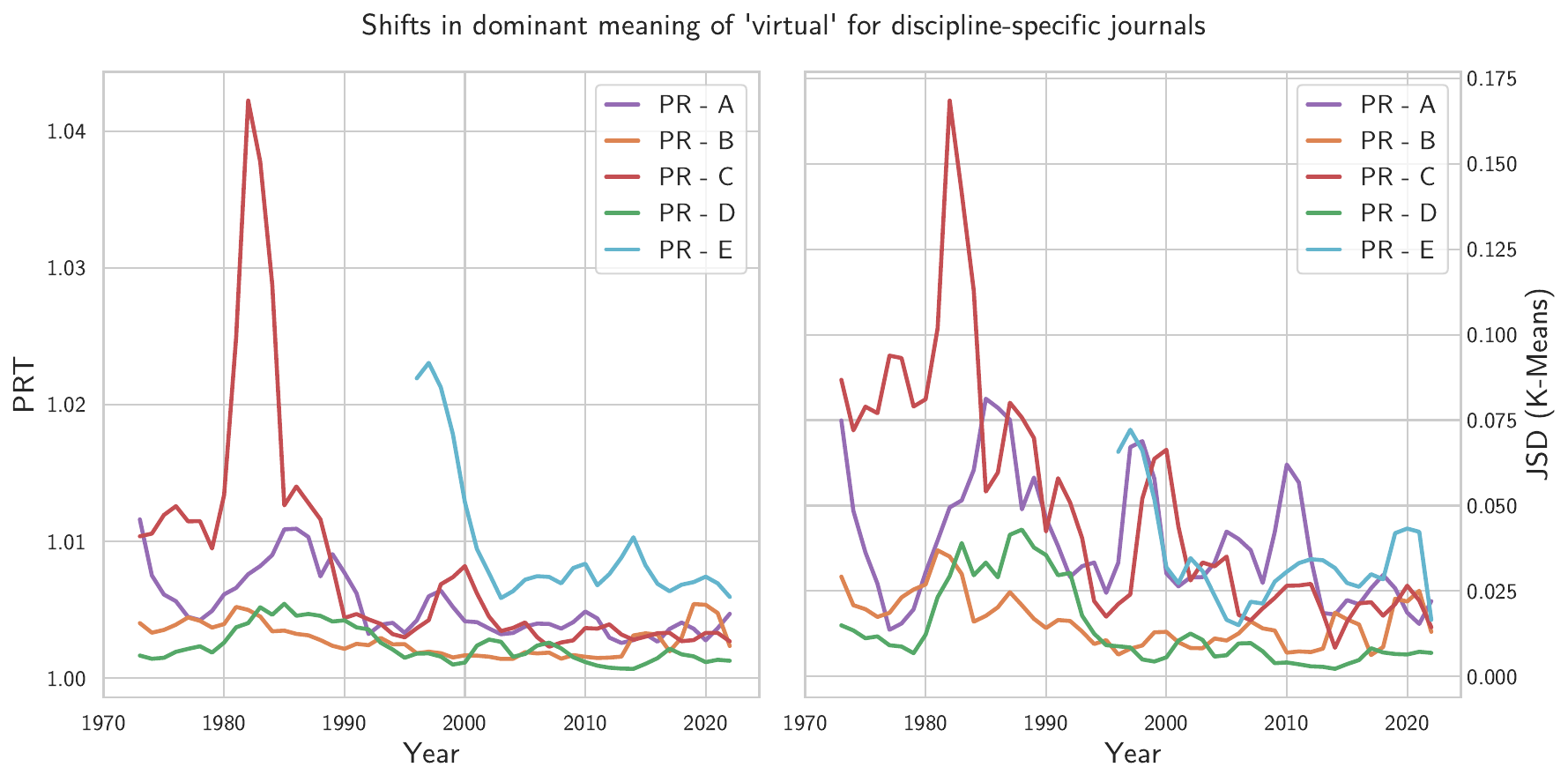}
  \caption{Shifts in dominant meaning in discipline-specific \textit{PR}-journals for \enquote{virtual}, using PRT (left) and JSD for K-Means clustering (right). For clarity, the rolling mean over 3 years is shown.}
  \label{fig:domMeaning_journals}
\end{figure}

\begin{figure}[h!]
  \centering
  \includegraphics[width=1\textwidth,center]{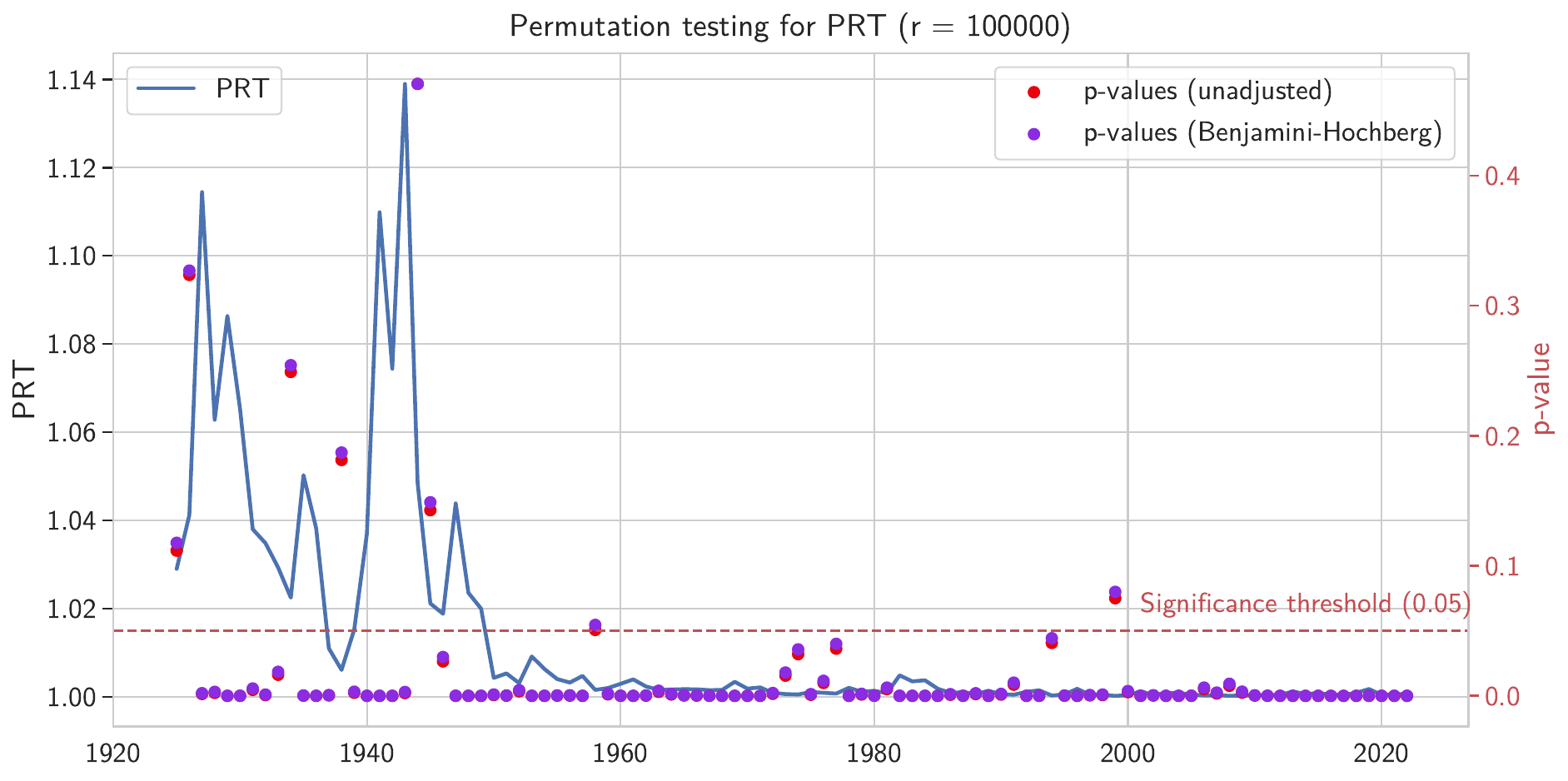}
  \caption{P-values (unadjusted and adjusted with Benjamini-Hochberg procedure) for the permutation-based statistical testing of the PRT-metric for \enquote{virtual}. The testing was done for 100.000 iterations (r). The dashed red line marks the significance threshold of 0.05.}
  \label{fig:prt_perm_testing}
\end{figure}

\begin{figure}[h!]
  \centering
  \includegraphics[width=1\textwidth,center]{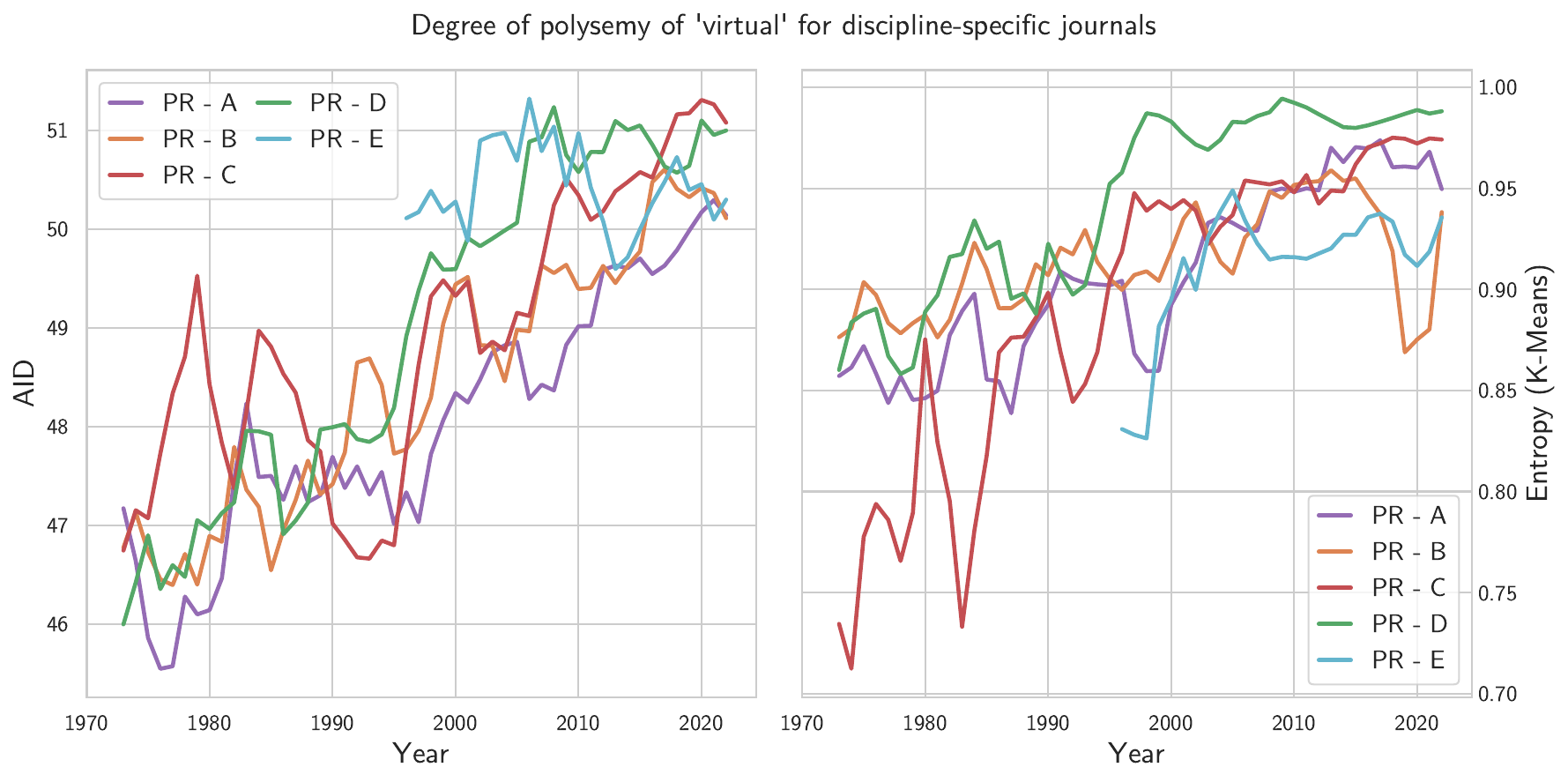}
  \caption{Changing degree in polysemy in discipline-specific \textit{PR}-journals for \enquote{virtual}, using AID (left) and normalized Shannon-Entropy for K-Means clustering (right). For clarity, the rolling mean over 3 years is shown.}
  \label{fig:polysemy_journals}
\end{figure}

\clearpage
\subsection{Tables}

\begin{table}[h!]
\caption{Top 5 lemmatized dependencies for \enquote{virtual} per discipline-specific journal. The number in brackets represents the share of the dependency per journal in all dependencies of the decade.}
\makebox[\textwidth][c]{
\begin{tabular}{|c|c|c|c|c|c|}
\hline
Top & PR - A & PR - B & PR - C & PR - D & PR - E \\
\hline
1 & state (12\%) & state (12\%) & photon (31\%) & photon (19\%) & particle (5\%) \\
2 & orbital (11\%) & transition (6\%) & state (12\%) & correction (8\%) & qubit (4\%) \\
3 & photon (10\%) & process (6\%) & excitation (4\%) & particle (3\%) & temperature (3\%) \\
4 & excitation (5\%) & excitation (5\%) & pion (2\%) & state (3\%) & time (2\%) \\
5 & transition (5\%) & approximation (4\%) & virtuality (2\%) & contribution (3\%) & point (2\%) \\
\hline
\end{tabular}
}
\label{table:deps_journals_virtual}
\end{table}

\begin{table}[h!]
\caption{Count of articles, \enquote{virtual}-embeddings and cleaned tokens in corpus per year for early-phase of analysis (up to 1950). After 1950 all three counts grow steadily, as can be seen in figure \ref{figure:corpus_overview}.}
\makebox[\textwidth][c]{
\begin{tabular}{|c|c|c|c|}
\hline
Year & Article count & \enquote{virtual}-embedding count & Cleaned tokens count \\ \hline
1924 & 3  & 11  & 6,715   \\ \hline
1925 & 7  & 14  & 10,723  \\ \hline
1926 & 4  & 4   & 6,600   \\ \hline
1927 & 4  & 24  & 5,343   \\ \hline
1928 & 3  & 7   & 4,660   \\ \hline
1929 & 2  & 23  & 4,384   \\ \hline
1930 & 7  & 13  & 24,682  \\ \hline
1931 & 9  & 32  & 32,688  \\ \hline
1932 & 9  & 19  & 15,540  \\ \hline
1933 & 8  & 13  & 32,767  \\ \hline
1934 & 9  & 15  & 31,648  \\ \hline
1935 & 18 & 53  & 26,422  \\ \hline
1936 & 26 & 65  & 84,642  \\ \hline
1937 & 36 & 128 & 114,524 \\ \hline
1938 & 16 & 47  & 33,787  \\ \hline
1939 & 29 & 93  & 51,193  \\ \hline
1940 & 10 & 47  & 16,443  \\ \hline
1941 & 11 & 31  & 26,949  \\ \hline
1942 & 8  & 24  & 7,643   \\ \hline
1943 & 3  & 4   & 3,628   \\ \hline
1944 & 6  & 14  & 4,512   \\ \hline
1945 & 12 & 30  & 53,587  \\ \hline
1946 & 12 & 26  & 28,033  \\ \hline
1947 & 20 & 68  & 35,555  \\ \hline
1948 & 34 & 107 & 43,308  \\ \hline
1949 & 78 & 208 & 135,181 \\ \hline
1950 & 62 & 170 & 119,892 \\ \hline
\end{tabular}
}
\label{table:corpus_counts}
\end{table}

\end{document}